\documentclass[letterpaper, 10 pt, journal]{IEEEtran}

\ifCLASSINFOpdf
  \usepackage[pdftex]{graphicx}
  \graphicspath{{figures/}}
\else
\fi

\usepackage{amsmath}
\usepackage{amssymb}

\ifCLASSOPTIONcompsoc
 \usepackage[caption=false,font=normalsize,labelfont=sf,textfont=sf]{subfig}
\else
 \usepackage[caption=false,font=footnotesize]{subfig}
\fi

\usepackage{url}
\usepackage{xcolor}

\usepackage{diagbox}

\def\shortName{CNEP}
\def\longName{Conditional Neural Expert Processes}
\def\otherShortName{CNMP}
\def\otherLongName{Conditional Neural Movement Primitives}

\begin{document}

\title{{\longName} for Learning Movement Primitives from Demonstration}

\author{Yigit YILDIRIM and Emre UGUR
\thanks{Yigit Yildirim and Emre Ugur are with Bogazici University. \texttt{yigit.yildirim@bogazici.edu.tr}}% <-this % stops a space
% \thanks{M. Shell was with the Department
% of Electrical and Computer Engineering, Georgia Institute of Technology, Atlanta,
% GA, 30332 USA e-mail: yigit.yildirim@boun.edu.tr}% <-this % stops a space
% \thanks{Manuscript received April 19, 2005; revised August 26, 2015.}}
}

%\markboth{Journal of \LaTeX\ Class Files,~Vol.~14, No.~8, August~2015}%
%{Shell \MakeLowercase{\textit{et al.}}: Bare Demo of IEEEtran.cls for IEEE Journals}

\maketitle

\begin{abstract}
Learning from Demonstration (LfD) is a widely used technique for skill acquisition in robotics. However, demonstrations of the same skill may exhibit significant variances, or learning systems may attempt to acquire different means of the same skill simultaneously, making it challenging to encode these motions into movement primitives. To address these challenges, we propose an LfD framework, namely the {\longName} ({\shortName}), that learns to assign demonstrations from different modes to distinct expert networks utilizing the inherent information within the latent space to match experts with the encoded representations. {\shortName} does not require supervision on which mode the trajectories belong to. 
% COM1: bunu atalim mi\yigit{Provided experiments on generated trajectories demonstrate the efficacy of CNEP. 
We compare the performance of {\shortName} against widely used and powerful LfD methods such as Gaussian Mixture Models, Probabilistic Movement Primitives, and Stable Movement Primitives and show that our method outperforms these baselines on multimodal trajectory datasets. The results reveal enhanced modeling performance for movement primitives, leading to the synthesis of trajectories that more accurately reflect those demonstrated by experts, particularly when the skill demonstrations include intersection points from various trajectories. We evaluated the CNEP model on two real-robot tasks, namely obstacle avoidance and pick-and-place tasks, that require the robot to learn multi-modal motion trajectories and execute the correct primitives given target environment conditions. We also showed that our system is capable of on-the-fly adaptation to environmental changes via an online conditioning mechanism. Lastly, we believe that {\shortName} offers improved explainability and interpretability by autonomously finding discrete behavior primitives and providing probability values about its expert selection decisions.

\end{abstract}

\begin{IEEEkeywords}
Learning from Demonstration, Deep Learning Methods
\end{IEEEkeywords}

% For peer-reviewed papers, you can put extra information on the cover
% page as needed:
% \ifCLASSOPTIONpeerreview
% \begin{center} \bfseries EDICS Category: 3-BBND \end{center}
% \fi
%
% For peer review papers, this IEEEtran command inserts a page break, and
% creates the second title. It will be ignored for other modes.
\IEEEpeerreviewmaketitle

\section{Introduction}
The capability of robots to comprehend and respond to dynamic environments is vital for their integration across various contexts. Specifically, most real-world tasks require the robots to model and construct spatiotemporal sensorimotor trajectories. Early applications involved manually recording the sensorimotor information generated by a demonstrator on a teleoperated robot and, later, autonomously following them on the robot \cite{segre1985explanation}. A skill was then represented by a series of primitive action segments selected according to simple conditional rules. These manually recorded controllers often fail outside the controlled environments due to the inherent characteristics of real-world environments as explained in \cite{ravichandar2020recent}.

Learning from Demonstration (LfD) is a widely adopted procedure in robotics that enables learning controllers to acquire new skills by observing an expert \cite{argall2009survey, kroemer2021review}. For this purpose, elaborate demonstrations of target skills are required in LfD. On the other hand, the set of expert demonstrations for a particular real-world skill may contain significant variances, or there might be multiple ways to achieve the same skill. These variances reflect the stochastic nature of the expert demonstrations, which poses the challenge of handling quantitatively and qualitatively different demonstrations for LfD-based skill-acquisition procedures.

Assume a human is given the task of teaching a robot a diverse set of skills, all initiated in the same environment setup. Given a large number of demonstrations without any further supervision about the type of motion, it is challenging to train a single system to handle such diversity. Instead, as Schaal et al. stated, ``the existence of movement primitives seems, so far, the only possibility how one could conceive that autonomous systems can cope with the complexity of motor control and motor learning'' \cite{schaal2005learning}. In this paper, we aim to develop a system that autonomously discovers the movement primitives while learning to generate the corresponding sensorimotor trajectories. 
For this, we introduce the {\longName} (CNEP) model, a generic and monolithic LfD framework to teach robots the necessary controllers to model and synthesize complex, multimodal sensorimotor trajectories. In previous approaches, such as \cite{schaal2005learning, paraschos2018using, garnelo2018conditional, seker2019conditional}, the multimodality aspect of target skills was not explicitly addressed as these approaches attempted to represent different modes with the same mechanism, leading to a seamless interpolation inside the demonstration space, which may lead to suboptimal behavior, as shown in \cite{pignat2019bayesian}. On the contrary, our {\shortName} is designed to model different modes in the demonstrations with different experts and generate the required motion trajectory by automatically selecting the corresponding expert. Our model is built on top of {\otherLongName} ({\otherShortName}) \cite{seker2019conditional}, which was shown to form robust representations to model complex motion trajectories from a few data points. CNMPs have an encoder-decoder structure that allows them to generate motion trajectories given a set of conditioning (observation) points. {\shortName} uses multiple decoders (experts) - instead of a single one - that are responsible for different modes in the given trajectories. Given the conditioning points and the output of the encoder network, a novel gating mechanism assigns probabilities to the experts, and the decoder with the highest probability is used to generate the motion trajectory from the encoded conditioning points. Besides the architectural contribution that includes the gating mechanism and multiple experts, we propose a novel loss function to ensure that all the experts are evenly utilized and an expert, when assigned, is selected with high probability.

We evaluated our system in different tasks that require learning different sets of movement trajectories, including arm and gripper trajectories, from a real robot system. We showed that {\shortName} outperforms the baseline models \cite{seker2019conditional,calinon2020mixture,paraschos2013probabilistic,perez2023stable} when the system is required to generate trajectories from common points of several demonstrations of increasing number of modes, and when generalizing into unseen conditioning points.

% COM1: deleted this paragraph. This paper unfolds as follows: Section \ref{sec:rel} reviews the related work in sensorimotor trajectory modeling, which enables modeling and synthesizing motion trajectories. Section \ref{sec:met} outlines the methodology of the {\shortName} procedure, explaining the structures and functions of the individual components. Section \ref{sec:exp} presents experimental results, establishing the effectiveness of {\shortName} in handling multimodal data. Finally, Section \ref{sec:con} offers concluding remarks, \yigit{limitations of the model,} and potential future directions.

\section{Related Work \label{sec:rel}}

%%%%%%%%%%%%%%%%%%%%%%%%%%%%%%%%%%%%%%%%
Equipping robots with the desired skills has been the driving force in robotics research. In initial studies, controllers with precise mathematical representations were used. These representations were formed using the physics-based dynamic models of the environment and the kinematic models of the agents \cite{canny1990exact}. Although accurate in controlled settings and computationally less intense, the applicability of the precise models was limited in realistic scenarios. This is mainly due to their constrained flexibility in the kinodynamic space of the system, preventing the generalization of acquired skills into novel conditions.

Movement Primitives (MP) formalism offers a compact and modular representation to create more flexible controllers.
% COM1: Bu cumleyi sildim. In the LfD setting, one of the most successful MP frameworks is the Dynamic Movement Primitives (DMP) }. 
For example, in DMPs \cite{schaal2005learning}, expert demonstration of a complex skill is encoded with a system of differential equations in the form of MPs. DMPs can be queried upon modeling to generate motion trajectories from start to end. Also, when integrated with closed-loop feedback, DMPs are proven suitable for real-time control as they adapt to changes and perturbations in real-time, offering robust performance in many applications \cite{saveriano2023dynamic}. However, only a single trajectory can be encoded by the classical DMP formulation, indicating that variabilities inside demonstrations are not considered. CNEPs, on the other hand, can encode distributions of trajectories.

Probabilistic approaches have been proposed to address the abovementioned requirements by offering flexible and robust modeling mechanisms. In this respect, Gaussian Mixture Models together with Gaussian Mixture Regression (GMM-GMR) and Hidden Markov Models (HMM) have been used in several studies \cite{zeestraten2017approach, pehlivan2015dynamic,girgin2018associative} to capture the variability of the task by learning the distributions of the demonstration data. The complexity of training and inference in HMMs increases as the dimensionality of the demonstrations increases, whereas variants of GMMs work well with high-dimensional data \cite{calinon2016tutorial}. Nonetheless, when the demonstration data of the task is sampled from a multimodal distribution, GMMs fail to select one of the modes. In contrast, state-transition probabilities of HMMs encode this information, enabling the synthesis of expert-like trajectories \cite{ugurcompliant}. The proposed {\shortName} addresses both of these issues. It utilizes multiple expert networks to handle multimodal data and can work with high-dimensional raw data coming directly from the sensors.

As stated in \cite{nguyen2009model, deisenroth2013gaussian}, the uncertainty in real-world tasks has been explicitly addressed using Gaussian Processes (GPs). As a result, the computational efficiency of learning adaptable and robust robotic controllers is improved to enable control in real-world tasks. Pure GP approaches are known to work well in Euclidean spaces. However, when the demonstration data displays non-Euclidean characteristics, such as rotation of robotic joints, further adjustments are required for GP methods \cite{arduengo2023gaussian}. This is not the case for {\shortName} as illustrated with the real robot tests where the complete trajectories in the non-Euclidean joint space are used as demonstrations.

% such as Gaussian Processes (GPs), Gaussian Mixture Models, and Hidden Markov Models \cite{nguyen2009model, williams2007modelling, calinon2016tutorial}

% to model the uncertainty
% with nondeterministic behaviors and interactions 

Addressing the in-task variability, Probabilistic Movement Primitives (ProMP) have been proposed to encode a distribution of trajectories \cite{paraschos2013probabilistic}. In \cite{paraschos2018using}, ProMPs were shown to provide improved generalization capabilities, enabling generated trajectories to be adapted so that they could pass through desired via points. However, ProMPs are composed of linear-Gaussian models that are limited to unimodal distributions and cannot represent multimodal datasets. Additionally, ProMPs cannot be efficiently trained or queried with high dimensional input.
%they fail when the demonstration set does not match a Gaussian distribution. Furthermore, the use of basis functions limits the applicability of this method to high-dimensional data.

%However, manually defined basis functions were used in ProMPs, making them unsuitable against high-dimensional data.

% In recent years, owing to the advances in machine learning, deep learning methods have been employed to offer alternatives to manual approaches. LSTMs have been successfully used in various studies to learn and generate multimodal trajectories \cite{alahi2016social}. Despite their success compared to the earlier methods, vast amounts of data are required to train LSTMs. Moreover, they are fed on their predictions to make further predictions, which induces an accumulated error problem in long-horizon predictions \cite{pekmezci2021learning}. Contrarily, {\shortName} can be queried to generate long-horizon trajectories without an error-accumulation issue.

As a deep LfD framework, {\otherLongName} ({\otherShortName}) is developed based on Conditional Neural Processes (CNP) \cite{garnelo2018conditional}, also aiming to handle high-dimensional sensorimotor data. {\otherShortName} can be used to learn movement primitives using sensorimotor data and construct trajectories that can be conditioned on real-time sensory data to enable real-time responses. It has been successfully applied to complex trajectory data across numerous studies and domains, such as \cite{yildirim2022learning, ada2023meta}. However, only a single query network is used in {\otherShortName} to decode different demonstrations of a movement primitive. As a result, trajectories are formed by interpolating between different modes of the same skill, which may lead to suboptimal results where the demonstration trajectories are multimodal or intersecting. 

Recently, the Stable Movement Primitives (Stable MP), \cite{perez2023stable}, is proposed to learn movement primitives, potentially belonging to multiple skills, using the same neural mechanism. 
% COM1: which same neural?
It offers the advantage of guaranteed precision at conditioning points. However, it requires supervision about the type of skill it learns or generates. Additionally, the system puts limitations on the conditioning mechanism.
% COM1: replaced with above Also, Stable MP cannot be conditioned on any other endpoint than the one it extracts from the demonstration data, hindering its generalizability. 
Contrarily, CNEP discovers trajectory types in an unsupervised manner and can be conditioned from any via point.

\section{Method \label{sec:met}}

\begin{figure*}[t]
    \begin{center}
        \includegraphics[width=0.99\textwidth]{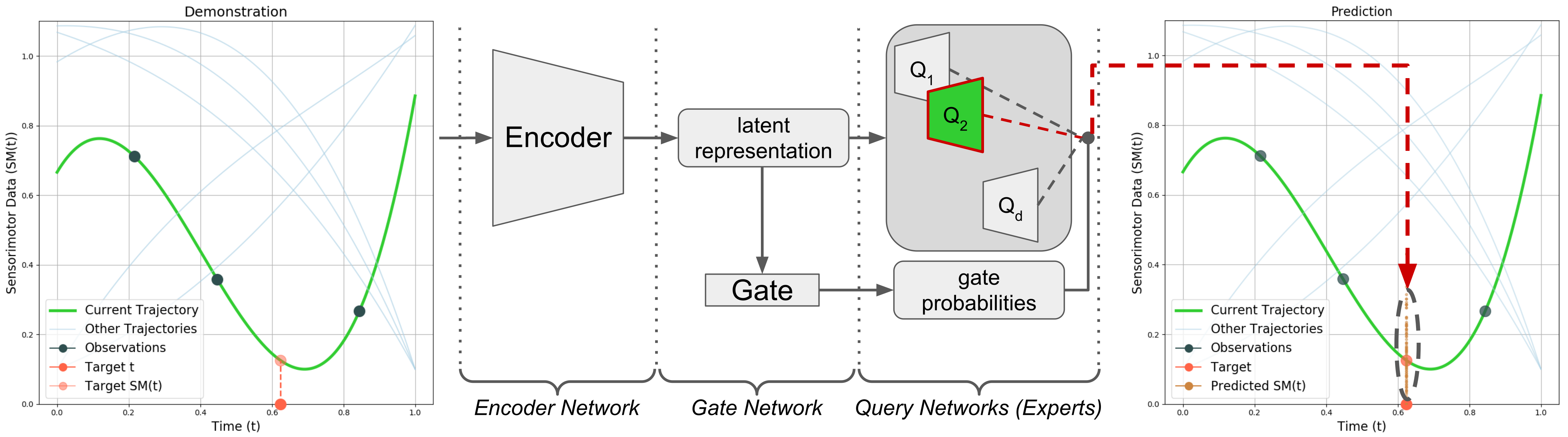}
        \caption{The CNEP model contains an Encoder, a Gate, and multiple Query Networks called experts. In this example, \textit{n}=3 observation points (shown with \textcolor{darkgray}{$\bullet$}) and \textit{m}=1 target timepoints (shown with \textcolor{red}{$\bullet$}) are randomly sampled on the input trajectory. The latent representation, the mean of $n$ observation encodings, is used by: (1) the Gate Network to find the responsible expert and (2) expert networks to generate their predictions as normal distributions at the target timepoint. Refer to Section \ref{sec:cnep_arch} for details. \label{fig:arc}}
    \end{center}
\end{figure*}

\subsection{Problem Formulation}
The skill-acquisition problem can be formulated as finding a sequence of motion commands that produce the desired movement \cite{gasparetto2007new}. Formally, the LfD system is expected to learn a function $\tau = f(t; X)$, where \textit{X} denotes specific criteria, such as the starting point at $t=0$ or the destination at any time $t$, using \textit{N} expert demonstrations, $D = \{\tau_1, \tau_2, \dotsc, \tau_N\}$. Despite the multimodality of the target skill, a resource-efficient solution with few demonstrations is also demanded to promote the applicability of the proposed approach in real-world settings where it is infeasible to provide so many demonstrations.

In this context, sensorimotor functions (\textit{SM(t)}) are utilized to refer to the temporal mapping of sensory inputs and motor outputs of a robot at time \textit{t}. Two important notions are encapsulated in the \textit{SM(t)} formalism: (1) how a robot senses its environment through sensors and (2) how it responds through actuators at any given moment. The perspective of representing complex skills as \textit{SM(t)} trajectories transforms skill-acquisition efforts into trajectory modeling and generation problems. A trajectory is formally defined as a temporal function, $\tau = \tau(t)$ following \cite{biagiotti2008trajectory}. Throughout this study, each trajectory $\tau$ is represented as an ordered list of sensorimotor values: $\tau = \{SM(t_1), SM(t_2), \dotsc, SM(t_T)\}$.

In the following section, after introducing the baseline ({\otherShortName}) method, we provide details of our proposed method ({\shortName}). As LfD frameworks, both are used to encode a set of trajectories from expert demonstrations. They take a set of observation (conditioning) points, in the form of (t, \textit{SM}(t)) tuples from a trajectory, and are expected to output the \textit{SM}$(t_q)$ value of any target timepoint, $t_q$. In practice, given varying observation points, the entire trajectory is generated by querying the system for all time points from $t_1$ to $t_T$.

\subsection{Background: CNMP}
{\otherShortName}, introduced in  \cite{seker2019conditional}, contains an Encoder and a Query Network. At the training time, a trajectory $\tau_i$ is first sampled from demonstrations, $D$. $n$ randomly sampled observation points from this trajectory are passed through the Encoder Network to generate corresponding latent representations. An averaging operation is applied to obtain a compact representation of the (n) input observations in the latent space. The average representation is then concatenated with $m$ random target timepoints and passed through the Query Network to output the distributions that describe the sensorimotor responses of the system at corresponding target timepoints. Here, $n$ and $m$ are random numbers, where $1 \leq n \leq n_{max}$ and $1 \leq m \leq m_{max}$. $n_{max}$ and $n_{max}$ are hyperparameters whose values are set empirically. The output is a multivariate normal distribution with parameters $(\mu_q, \Sigma_q)$. The loss is calculated as the negative log-likelihood of the ground-truth value under the predicted distribution as follows:

\begin{equation}
\label{eq:loss_cnmp}
\mathcal{L} = -\: log\: P(\:\textit{SM}(t_q)\: |\: \mathcal{N}(\mu_q,\: softplus(\Sigma_q)\:)\:).
\end{equation}

\noindent This loss is backpropagated, updating the weights of both the Encoder and the Query networks. More details can be found in \cite{seker2019conditional}.

\subsection{Proposed Approach: {\shortName}}
\subsubsection{Architecture Overview \label{sec:cnep_arch}}

The proposed architecture and the workflow are illustrated in Fig. \ref{fig:arc}. Similar to the {\otherShortName}, the input to the system is composed of $n$ observations, which are passed through the Encoder Network to generate the latent representation. The latent representation is fed into our novel Gate Network to produce the gate probabilities for all experts. Simultaneously, they are concatenated with $m$ target timepoints and are passed through all experts to generate $SM$ predictions at target timepoints. During training, the Gate Network's output is combined with the experts' outputs to compute the overall loss. After training, when the system is asked to generate a response for a target, only the prediction of the expert with the highest gate probability is outputted. An example case is shown in Fig. \ref{fig:arc}, where after producing the latent representation for n=3 observation points, the gating mechanism outputs a relatively high probability for the second expert, appointing it as the responsible expert for this query. Therefore, its response for m=1 target timepoint is selected as the output of the entire system. The system is trained end-to-end, as detailed in the next section.

\subsubsection{Training Procedure \label{sec:cnep_train}}

\begin{figure*}[t]
    \begin{center}
        \includegraphics[width=0.99\textwidth]{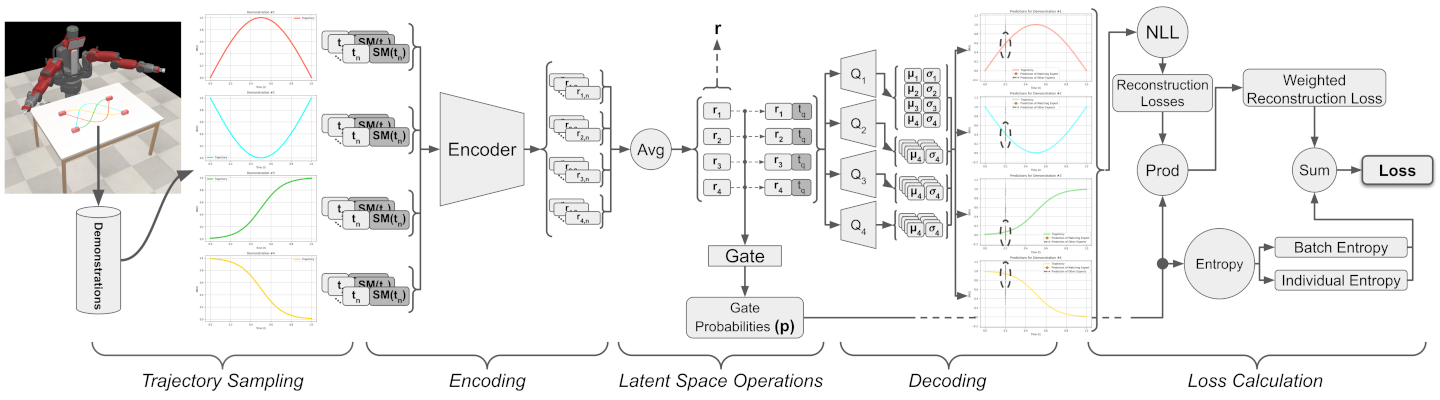}
        \caption{Initially, observation points from the input trajectory are mapped to the latent space by the Encoder Network. The averaged representations ($\mathbf{r}$) are (1) fed into the Gate Network and (2) concatenated with target points ($\mathbf{r_q}$) and fed into the Query Networks. While the candidate predictions are outputted by all Query Networks, the probabilities for each trajectory-expert pair ($\mathbf{p}$) are generated by the Gate Network. These probabilities are used in calculating the values of loss components as explained in Section \ref{sec:cnep_train}.  \label{fig:cnep}}
        
    \end{center}
\end{figure*}

The training phase of the {\shortName} model is depicted in Fig. \ref{fig:cnep}. From a randomly selected trajectory, $\tau_i$, $n$ observation points are randomly sampled to form $(t, \textit{SM}(t))$ tuples and given to the Encoder Network. A compact representation estimate for $\tau_i$, $\mathbf{r_i}$, is obtained by averaging the output of the Encoder Network for the $n$ conditioning points. In Fig. \ref{fig:cnep}, parallel processing of a batch of trajectories is shown.

\paragraph{Calculating Gate Probabilities} The gate probability is the probability that the set of conditioning points and their underlying trajectory will be generated by the corresponding expert, as predicted by our system. For each expert $e$, the gate probability $p_{i,e}$ is calculated by the Gate Network, which takes as input the average latent activation ($\mathbf{r_i}$) of the corresponding conditioning points and passes it through a linear layer followed by the Softmax function. The number of experts, $d$, is a hyperparameter in our model.

\paragraph{The Reconstruction Loss}
Each expert predicts a normal distribution for the \textit{SM} values at target timepoints, and a reconstruction loss is calculated using the ground-truth \textit{SM} values. For this, $m$ target timepoints ($\mathbf{t_q}$) are randomly sampled, where m is set to 1 in Fig. \ref{fig:cnep} for simplicity.
An $(\mathbf{r_i}, \mathbf{t_q})$ tuple is passed through all experts ($Q_e$s) to generate their predictions. The negative log-likelihood of the actual $SM(\mathbf{t_q})$ under predicted distribution is computed as the \textit{reconstruction loss} of the corresponding expert as follows:
\begin{align*}
L_{e} = -\: log\: P(\:\textit{SM}(\mathbf{t_q})\: |\: \mathcal{N}(\mathbf{\mu_e},\: softplus(\mathbf{\Sigma_e})\:)\:) 
\end{align*}

\noindent where $\mathbf{\mu_e}$ and $\mathbf{\Sigma_e}$ are the outputs of the Query Network, $Q_e$.

Next, the combined reconstruction loss for the trajectory $\tau_i$ is calculated by taking the weighted sum of $L_{e}$s of all experts:

\begin{align*}
\mathcal{L}_{rec}^{i} = \frac{1}{d} \sum_{e=1}^{d} L_{e} \times p_{i, e}
\end{align*}

\noindent where $p_{i, e}$ is the probability of expert $e$ for the latent representation $\mathbf{r_i}$. For a batch of trajectories (Fig. \ref{fig:cnep}), the \textit{weighted reconstruction loss} $\mathcal{L}_{rec}$ is the mean $\mathcal{L}_{rec}^{i}$, where $1 \leq i \leq b$, and \textit{b} is the batch size.
% Initially, a batch of trajectories is sampled from the set of all demonstrations, $B = \{\tau_1, \tau_2, \dotsc, \tau_b\}$. Then, $n<n_{max}$ stochastically sampled observation points are selected from each trajectory to form the input to the Encoder Network, forming a $b\:\times\:n$ matrix of $(t, \textit{SM}(t))$ tuples as input to the model. The Encoder Network produces $b\:\times\:n$ encodings, which are then averaged to generate $b$ latent representations, $r \in \mathbb{R}^{bxh}$, one for each $\tau$ in the batch.

\paragraph{Expert Assignment Losses} We would like our system to avoid selecting the same expert for all possible modes. For this purpose, the entropy of the expert activation frequencies over a batch of training trajectories should be maximized. We use \textit{batch entropy} ($\mathcal{L}_{\textrm{batch}}$) to enforce this constraint. In the meantime, we would like the system to attribute a high probability when assigning an $(\mathbf{r_i}, \mathbf{t_q})$ tuple to one of the experts. For this, the entropy of the gate probabilities, $p_i$, should be minimized. We use \textit{individual entropy} ($\mathcal{L}_{ind}$) to enforce this constraint.
%In other words, to prevent overutilization of the same expert and to promote expert specialization, the mean expert activation probabilities are expected to approximate a uniform distribution as the training progresses.

Formally, $\mathcal{L}_{\textrm{batch}} \in \mathbb{R}$ is calculated as follows:

\begin{align*}
\mathcal{L}_{\textrm{batch}} = -\sum_{e=1}^{d} \left( \frac{1}{b} \sum_{i=1}^{b} p_{i, e} \right) \log\left( \frac{1}{b} \sum_{i=1}^{b} p_{i, e} \right)
\end{align*}

\noindent where, again, $b$ is batch size, and $d$ is the number of experts.

Subsequently, $\mathcal{L}_{ind} \in \mathbb{R}$ is computed as follows:

\begin{align*}
&\mathcal{L}_{ind} = \frac{1}{b} \sum_{i=1}^{b} \left( -\sum_{e=1}^{d} p_{i, e} \log(p_{i, e}) \right)
\end{align*}

The entire system is trained in an end-to-end manner where the parameters of the Encoder, the Gate, and the Query Networks are trained simultaneously in a supervised way. The overall loss function is a linear combination of the abovementioned three components: (1) the \textit{weighted reconstruction loss}, (2) batch-wise expert activation loss, the \textit{batch entropy}, and 3) trajectory-wise expert selection probability, the \textit{individual entropy}.
The dynamic nature of expert selection necessitates careful handling during training to achieve the right balance between expert adaptation and overall system stability. Essentially, as the model attempts to decrease the reconstruction loss, it stimulates experts to make accurate predictions. Moreover, while the model attempts to increase batch entropy, it promotes expert specialization by preventing the overutilization of any expert. Lastly, the individual entropy component of the loss indirectly implies the confidence of latent representation-expert matching. As the model attempts to decrease this value, it contributes to the specialization of experts by assigning similar representations to the same expert. As a result, the following overall loss function is used:
\begin{equation}
\label{eq:cnep_loss}
\mathcal{L}_{total} = \alpha_1 \times \mathcal{L}_{rec} + \alpha_2 \times \mathcal{L}_{batch} + \alpha_3 \times \mathcal{L}_{ind},
\end{equation}

\noindent where weighting coefficients of these components, $\alpha_1$, $\alpha_2$, and $\alpha_3$, are found empirically by the grid search technique using a specific library, called Weights \& Biases, \cite{biewald2020experiment}.

\subsubsection{PID Controller \label{sec:pid_control}}

To execute learned skills on the robot, a PID controller is appended to the end of our system, ensuring that synthesized trajectories pass through specified observation points. The predicted SM trajectories are fed into this controller prior to the execution. The control signal \( u(t) \) for each timestep is calculated using the PID formula:
\[
\mathbf{u}(t) = K_p \cdot \mathbf{e}(t) + K_i \cdot \int \mathbf{e}(t) \, dt + K_d \cdot \frac{d}{dt} \mathbf{e}(t)
\]

\noindent where \( \mathbf{e}(t) \) is the error vector at time \( t \), and \( K_p \), \( K_i \), and \( K_d \) are the proportional, integral, and derivative gains, respectively, which define the controller's behavior.

The error \( \mathbf{e}(t) \) is the difference between the current point and the conditioning point. A decay mechanism of certain timesteps is incorporated into the error calculation. This ensures that the corrections diminish, allowing the trajectory to converge smoothly toward the desired path.

\section{Experiments and Results \label{sec:exp}}

This section presents evaluations of the proposed model by showing its performance in learning movement primitives from artificial and real-world demonstrations, shows the influence of the system components via ablation studies, and compares it with several baseline MPs.
% COM1: replaced with above.
%through multiple experiments. The first set of tests uses synthetically produced trajectories as sensorimotor demonstrations and assesses the model against an ensemble of movement primitive-based approaches. Furthermore, two ablation studies exhibit the effects of individual components of the model. Finally, the proposed approach is evaluated on two tasks with a real robotic manipulator. These real-world experiments demonstrate the potential of the CNEP model for practical cases.}

\subsection{Modelling Different MPs with Common Points}
\begin{figure}[t]
    \centering
    \subfloat[\label{fig:mixed_obs}]{\includegraphics[width=0.37\columnwidth]{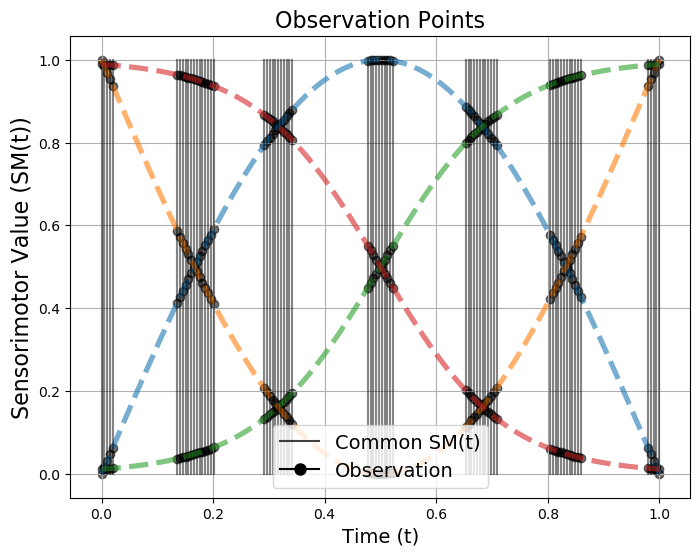}}
    \enspace
    \subfloat[\label{fig:comb_ex}]{\includegraphics[width=0.6\columnwidth]{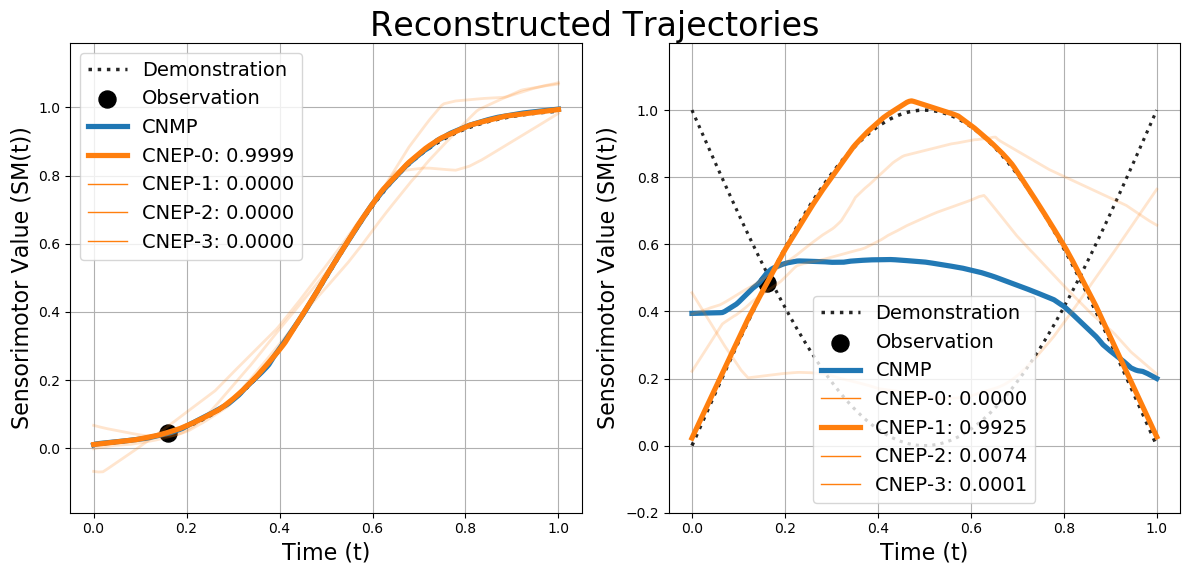}}
    \caption{(a) SM trajectories and observation points used in the comparison. (b) Conditioned from different points. While CNMP might produce an average response, CNEP successfully generates the target trajectory. \label{fig:mixed_res}}
    \enspace
\end{figure}

%When showing skill with diverse demonstrations, the corresponding trajectories probably pass through the same points. For example, consider a scenario where the arm is required to execute different point-to-point movements while passing through common apertures between obstacles. It is challenging for LfD systems to generate correct SM values when only conditioned from these intersecting points. 
In this section, we aim to evaluate the performance of our model in a scenario where the model needs to learn from a diverse set of demonstrations with common points.
%and compare it with the baseline when distinct trajectories with several common points are required to be learned. 
Both {\shortName} and CNMP are trained with the dataset of four SM trajectories shown in Fig. \ref{fig:mixed_obs}, where the trajectories intersect at various points. 
%Here, we aim to evaluate the performance of our model when it is required to generate the trajectory given observation points close to the common points.
Fig. \ref{fig:comb_ex} provides sample trajectory generations from CNEP and CNMP models. The numbers on the legend next to {\shortName} correspond to the probability of assignment of one of the experts. The dashed lines correspond to the demonstration trajectories in the dataset. When these models are conditioned from points unique to single trajectories, both models generate the required trajectories successfully (Fig. \ref{fig:comb_ex}-bottom). However, when they are conditioned from points close to the intersection, CNMP starts failing, whereas {\shortName} successfully generates the correct trajectories, as shown in the top row of Fig. \ref{fig:comb_ex}. As shown, while the CNMP model generates a trajectory that resembles an interpolated trajectory, our model assigns a high probability to one of the experts, which generates a trajectory close to one of the demonstrations in the dataset. Note that the obtained high probability value, when the conditioning point is very close to the intersection, is due to the individual entropy term in our loss function and our winner-take-all strategy.

\subsection{Comparison on Trajectories with Increasing Complexities}

\begin{figure}[t]
    \centering
    \subfloat{\includegraphics[width=0.49\columnwidth]{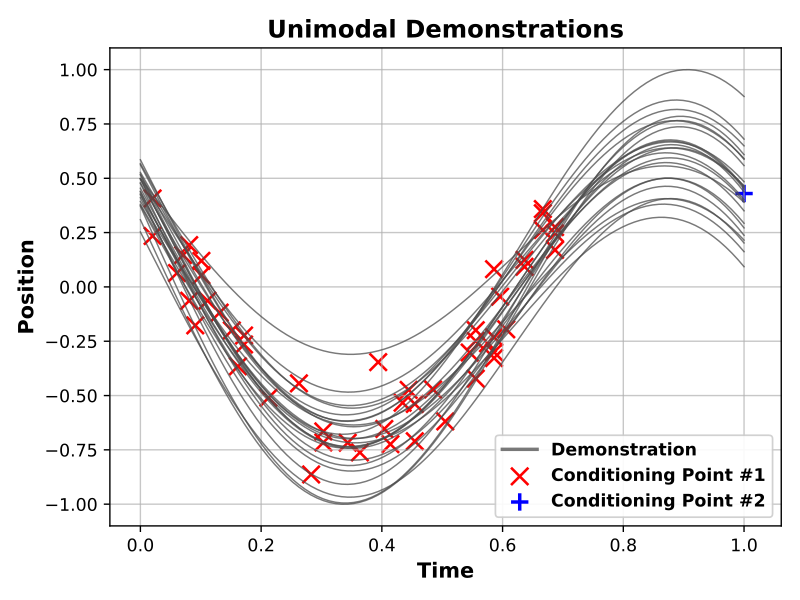}}
    \enspace
    \subfloat{\includegraphics[width=0.49\columnwidth]{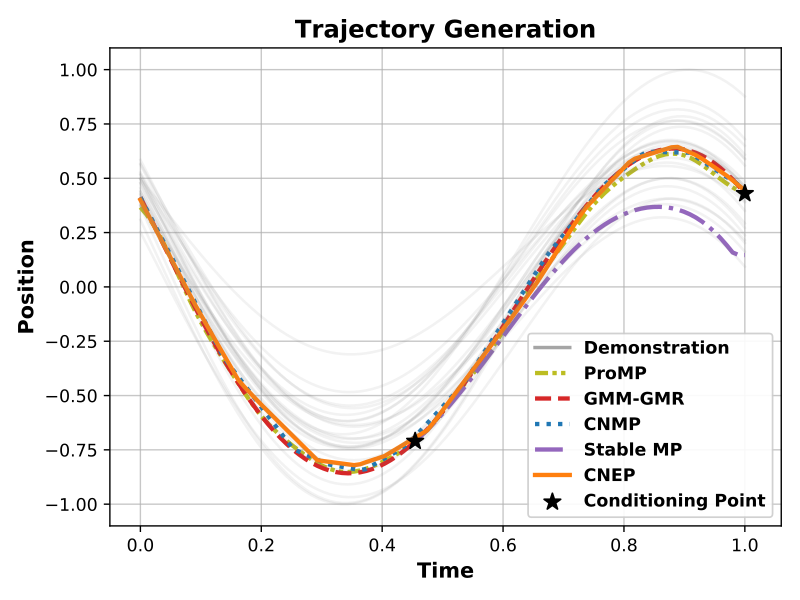}}\\
    \subfloat{\includegraphics[width=0.49\columnwidth]{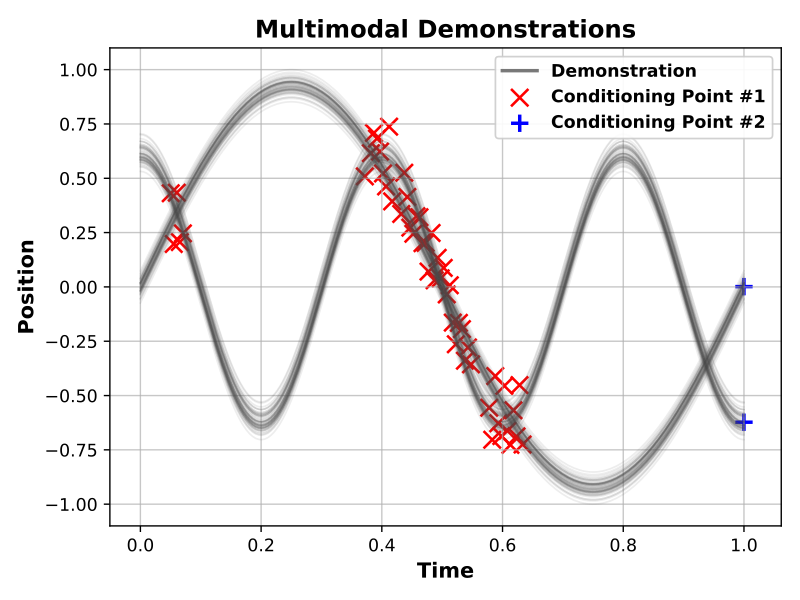}}
    \enspace
    \subfloat{\includegraphics[width=0.49\columnwidth]{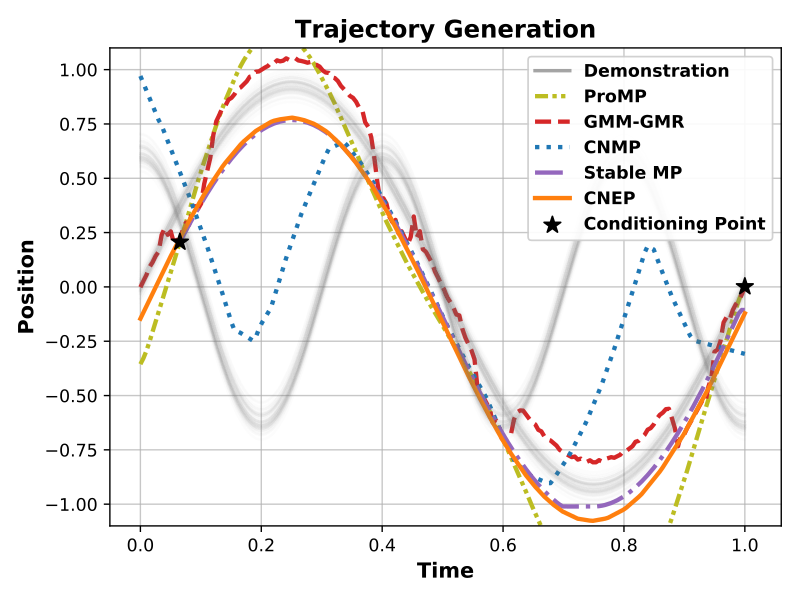}}\\
    \subfloat{\includegraphics[width=0.49\columnwidth]{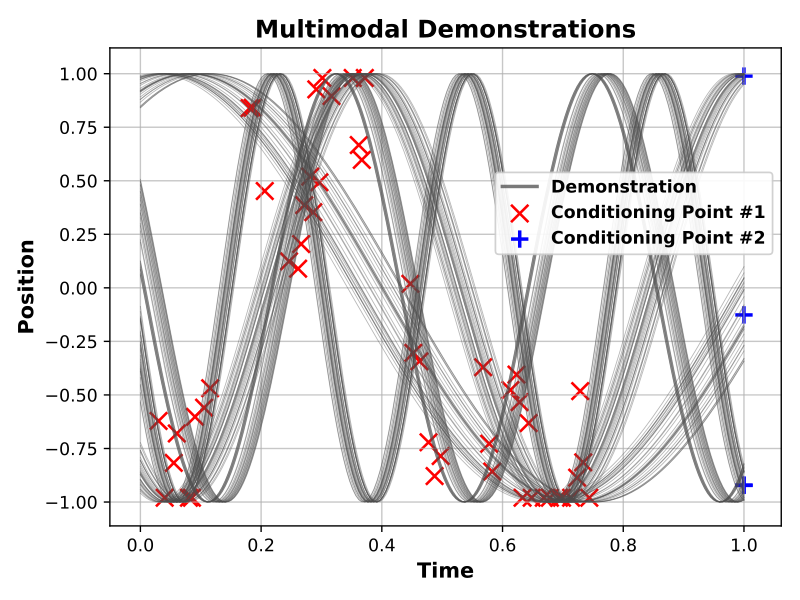}}
    \enspace
    \subfloat{\includegraphics[width=0.49\columnwidth]{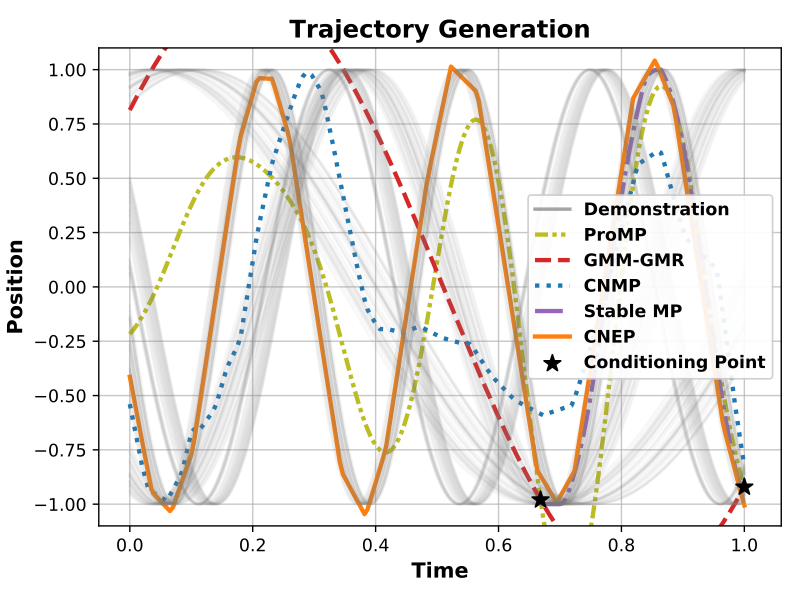}}
    \caption{The left column presents datasets of sensorimotor trajectories with increasing complexities. Correspondingly, the right column presents synthesized trajectories on an example run upon training. Modeling the skills these trajectories realize becomes more challenging as the number of modalities increases. However, different experts inside the CNEP model can successfully handle the increasing complexity. \label{fig:synth_trajs}}
    
\end{figure}

To present the advantages of the CNEP model, its performance is evaluated and compared to a group of baseline methods, including ProMP, GMM-GMR, CNMP, and Stable MP.
%\footnote{\yigit{\cite{fabisch2024movement_primitives} and \cite{fabisch2021gmr} were used as ProMP and GMM implementations.}}. 
Three datasets of sensorimotor trajectories with gradually increasing complexities are created. These three datasets are shown on the left side of Fig. \ref{fig:synth_trajs}. 
After training each method with the datasets, a test set is created with 50 pairs of intermediate and end conditioning points and their corresponding ground-truth trajectories. The intermediate conditioning points were sampled from the regions where trajectories of different modes come close, as shown with the red cross (x) markers. The endpoints were sampled from the training range, as shown with the blue plus (+) markers in the same figure.
The ground truth trajectories in evaluations were selected as the trajectories closest to the conditioning points in the demonstration set. The Mean-Squared Error (MSE) along the generated and ground truth trajectories were calculated for each query, and the mean and standard deviation of the errors are reported in Table \ref{tab:comp}.
Each method successfully generated plausible trajectories with low errors when the demonstration trajectories come from a unimodal distribution. However, as the complexity of the dataset increased, our method {\shortName} outperformed all other methods. The right side of Fig. \ref{fig:synth_trajs} features several sample trajectories generated by these methods. As shown, while CNMP, GMM-GMR, and ProMP generate extrapolated, interpolated, or shifted trajectories that might correspond to mixed combinations of the demonstrated ones, our CNEP model can select the correct primitive and generate the target trajectory successfully.

\begin{table}[h]
\centering
\caption{MSE over generated trajectories when conditioned from intermediate points \label{tab:comp}}
\begin{tabular}{|l|c|c|c|}
\hline
\textbf{} & \textbf{Unimodal} & \textbf{Bimodal} & \textbf{Multimodal} \\ \hline
\textbf{ProMP}       & 0.017 $\pm$ 0.010 & 0.312 $\pm$ 0.314 & 0.199 $\pm$ 0.191  \\ \hline
\textbf{GMM-GMR}     & 0.017 $\pm$ 0.009 & 0.096 $\pm$ 0.120 & 0.162 $\pm$ 0.065  \\ \hline
\textbf{CNMP}        & 0.015 $\pm$ 0.006 & 0.043 $\pm$ 0.078 & 0.112 $\pm$ 0.074  \\ \hline
\textbf{Stable MP}  & 0.016 $\pm$ 0.014 & 0.017 $\pm$ 0.035 & 0.058 $\pm$ 0.071 \\ \hline
\textbf{CNEP}       & \textbf{0.015 $\pm$ 0.005} & \textbf{0.013 $\pm$ 0.013} & \textbf{0.027 $\pm$ 0.042}  \\ \hline
\end{tabular}
\end{table}

\subsubsection{Influence of the Loss Components}

The loss term is one of the novel contributions of this paper, and the influence of the components in the loss term requires further investigation. For this, three variants of the CNEP model were created: 
\begin{enumerate}
    \item CNEP-Uniform (CNEP-Uni): 
    CNEP model where the coefficient of the batch entropy term, $\alpha_2$ in Eq. \ref{eq:cnep_loss}, is set to 0. The batch entropy term promotes using all decoders to learn different skills collectively.
    \item CNEP-UnSpecialized (CNEP-UnSpec): 
    CNEP model where the coefficient of the individual entropy term, $\alpha_3$ in Eq. \ref{eq:cnep_loss}, is set to 0. The individual entropy term promotes decoder specialization by rewarding high-probability matches between experts and encoded representations.
    \item CNEP-Reconstruction-only (CNEP-Rec): 
    CNEP model where coefficients of both entropy-related loss terms, $\alpha_2$ and $\alpha_3$ in Eq. \ref{eq:cnep_loss}, are set to 0. This model only considers the reconstruction loss.
\end{enumerate}

In addition, our CNEP model follows a winner-take-all approach: It outputs only the prediction of the responsible expert. Another approach might be using a weighted mixture of all experts (MoE). To justify our decision and compare it with its MoE variant, a CNEP model with a Mixture-of-Experts (CNEP-MoE) has been created. These 4 variants were trained alongside the original CNEP on the datasets shown in Fig. \ref{fig:synth_trajs}. Table \ref{tab:abl0} presents the MSE errors computed along the predicted and ground-truth trajectories. As shown, using all loss terms and our winner-take-all approach outperformed its variants.

\begin{table}[t]

\centering
\caption{MSE over generated trajectories}\label{tab:abl0}
\begin{tabular}{|l|c|c|c|}
\hline
\textbf{} & \textbf{Unimodal} & \textbf{Bimodal} & \textbf{Multimodal} \\ 
& & & \textbf{(4 modes)} \\ \hline
\textbf{CNEP-Uni}    & 0.042 $\pm$ 0.042 & 0.185 $\pm$ 0.057 & 0.120 $\pm$ 0.079 \\ \hline
\textbf{CNEP-UnSpec}    & 0.033 $\pm$ 0.031 & 0.069 $\pm$ 0.096 & 0.050 $\pm$ 0.077 \\ \hline
\textbf{CNEP-Rec}    & \textbf{0.031 $\pm$ 0.028} & 0.113 $\pm$ 0.045 & 0.064 $\pm$ 0.069 \\ \hline
\textbf{CNEP-MoE}   & 0.046 $\pm$ 0.049 & 0.061 $\pm$ 0.071 & 0.040 $\pm$ 0.061 \\ \hline
\textbf{CNEP}       & 0.046 $\pm$ 0.049 & \textbf{0.058 $\pm$ 0.071} & \textbf{0.028 $\pm$ 0.049} \\ \hline
\end{tabular}
\end{table}

\subsubsection{Influence of the Number of Experts}

The number of specialized experts inside the CNEP model also affects the overall performance. Results gathered in Table \ref{tab:comp3} compare 3 CNEP models with 2, 4, and 8 experts. Increasing the number of experts in the CNEP model to a number greater than or equal to the number of modalities of the demonstration data improves CNEP's performance. For example, when the data comes from a 2-modal distribution, the difference between CNEP-2 and CNEP-8 is negligible. As the number of modalities increases, the model must use more experts.
% COM1: replaced the below with above
%the difference between these models becomes more pronounced.}

\begin{table}[h]
\caption{MSE over generated trajectories
\label{tab:comp3}}
\centering
\begin{tabular}{|l|c|c|c|}
\hline
\textbf{} & \textbf{2 Modes} & \textbf{4 Modes} & \textbf{6 Modes} \\ \hline
\textbf{CNEP-2} & 0.005 $\pm$ 0.087 & 0.045 $\pm$ 0.038 & 0.063 $\pm$ 0.167 \\ \hline
\textbf{CNEP-4} & 0.005 $\pm$ 0.090 & 0.042 $\pm$ 0.038 & 0.033 $\pm$ 0.148 \\ \hline
\textbf{CNEP-8} & 0.004 $\pm$ 0.048 & 0.042 $\pm$ 0.04 & 0.032 $\pm$ 0.168 \\ \hline
\end{tabular}
\end{table}

%An illuminating example is presented in Fig. \ref{fig:comb_ex}. In the second row of this plot, trajectories produced by {\otherShortName} are seen to pass between two candidate trajectories. This is not the case for {\shortName} as one of the candidate trajectories is picked by this model. 
% The above examples show how models behave over exemplary cases. When all the points in Fig. \ref{fig:mixed_obs} are used, it can be seen in Fig. \ref{fig:comb_res} that fewer errors are produced by {\shortName}.
%due to this characteristic. The next section presents real-world tests to justify why the tendency toward one of the modalities is believed to be crucial in realistic LfD approaches, according to the authors.

\subsection{Learning from Real Robot Demonstrations \label{sec:real_world}}

In this section, we evaluate the performance of our system on two real-world tasks using a robotic manipulator, the Baxter robotic platform \cite{cremer2016performance}. In both experiments, demonstrations were collected following the kinesthetic teaching approach \cite{argall2009survey}. The first experiment demonstrates the learning and generalization capabilities of the CNEP model, where the robot is expected to realize an obstacle avoidance task. The second experiment illustrates how CNEP performs against high-dimensional sensory data on a more complex task that requires picking and placing objects in different configurations.

\subsubsection{Obstacle Avoidance}
\begin{figure}[t]
    \centering
    \subfloat{\includegraphics[width=0.32\columnwidth]{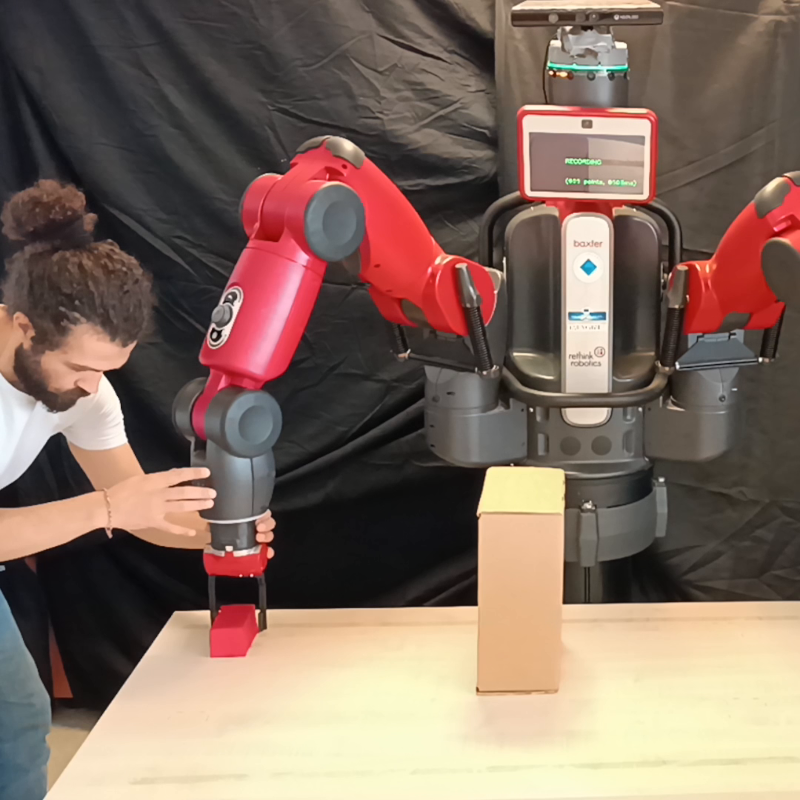}}
    \enspace
    \subfloat{\includegraphics[width=0.32\columnwidth]{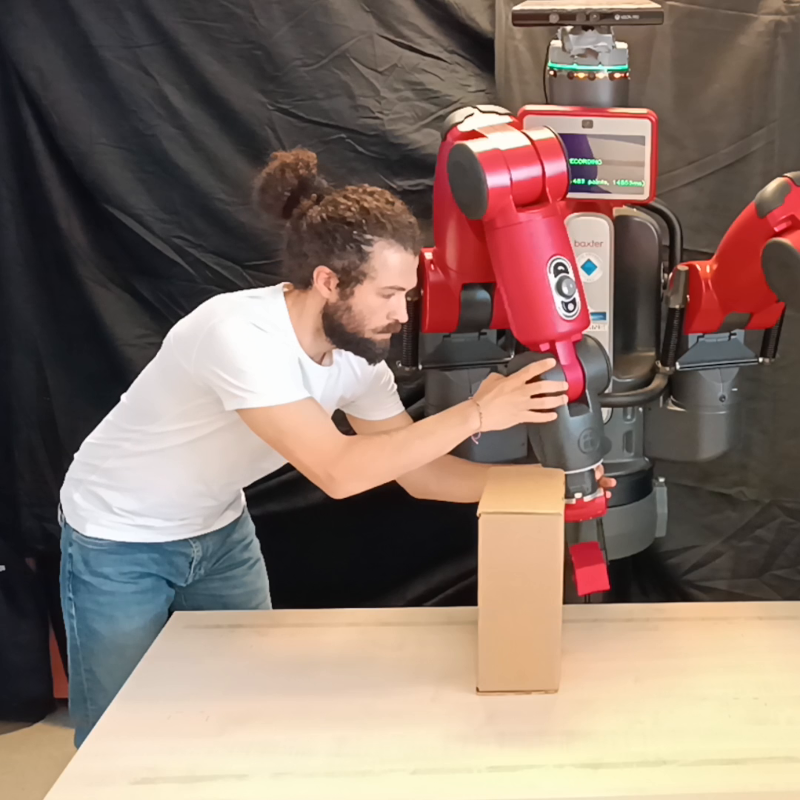}}
    \enspace
    \subfloat{\includegraphics[width=0.32\columnwidth]{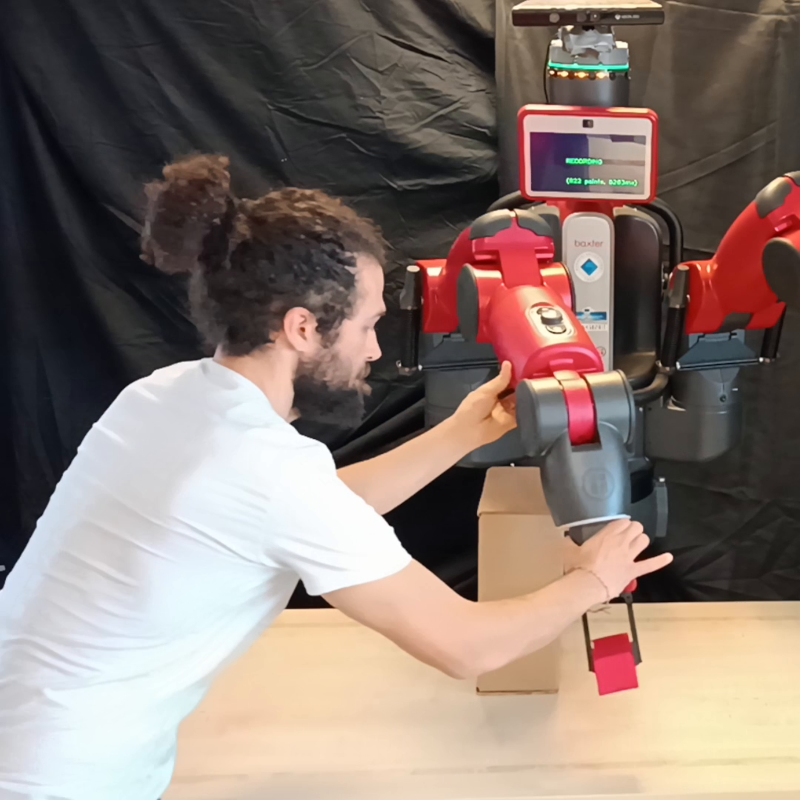}}
    \caption{Demonstrations of the obstacle avoidance skill are being performed by an expert. Kinesthetic teaching is used to generate sensorimotor demonstrations. Later, this data is used to train {\shortName} and {\otherShortName} models. \label{fig:baxter_cs}}    
\end{figure}

The first experiment investigates the advantages of CNEP over CNMP in a multimodal obstacle avoidance task where two demonstrations that avoid obstacles from different sides were shown by an expert. The SM demonstrations include (1) the timestamped points of seven joints of the manipulator in the joint space and (2) the 7-dimensional pose of the end effector in the Cartesian space (Fig. \ref{fig:baxter_cs}). We carried out evaluations both in Cartesian and joint spaces.
% COM1: remove First, we trained both models on the data collected in the Cartesian space, and for the second one, we used the data collected in the joint space. 
To investigate the capabilities of the CNEP model, we chose to condition both the {\otherShortName} and {\shortName} on distinctive starting points. These conditioning points were selected to lie at the midpoint between the initiation points of the two demonstrations to compare the generalization capabilities offered by the models.

\begin{figure}[t]
    \centering
    \subfloat{\includegraphics[width=\columnwidth]{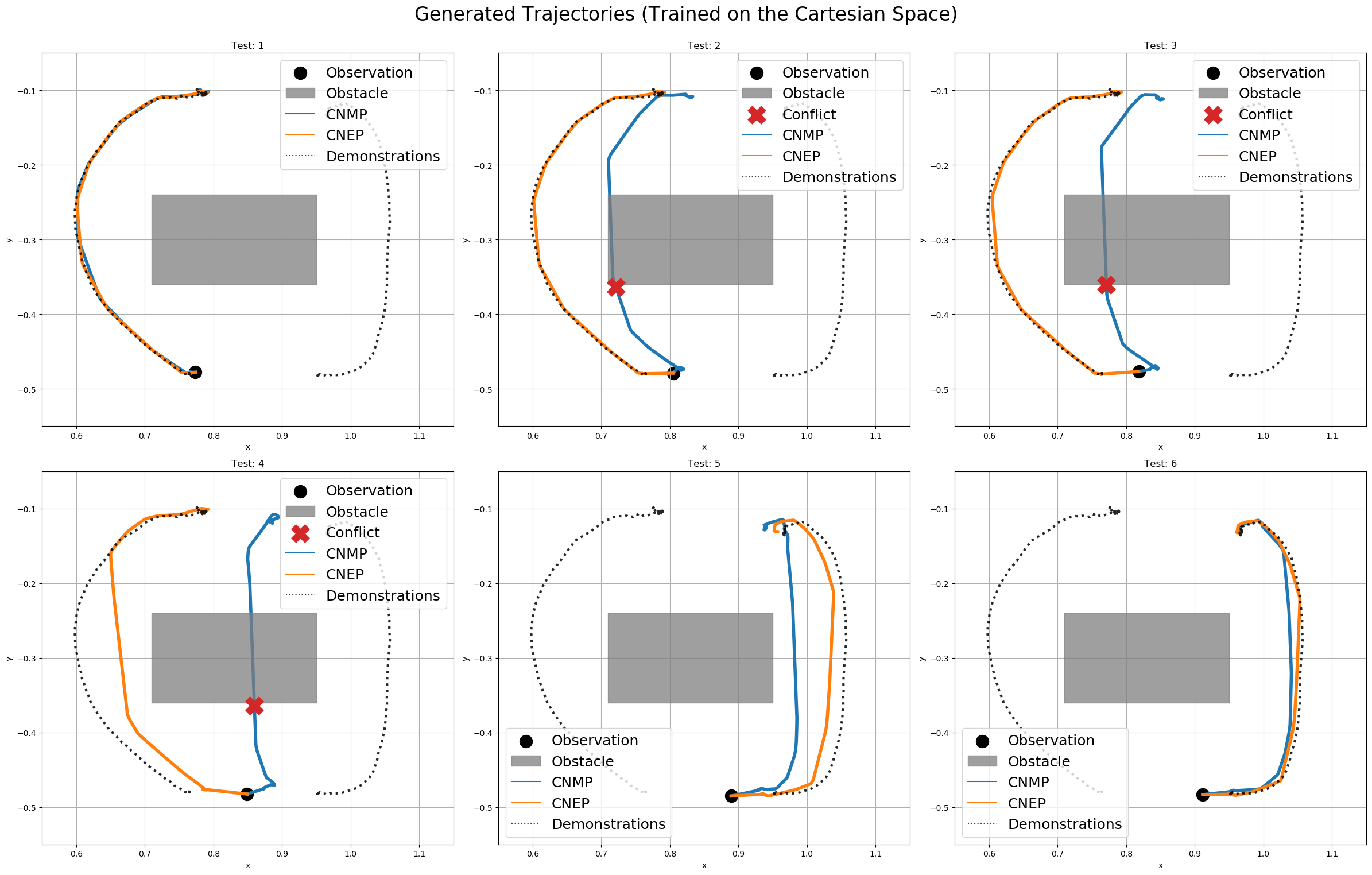}}
    \caption{Using the expert demonstrations (shown with dashed lines), two comparisons are made: one in the Cartesian space and another in the joint space. Here, only the former comparison is presented for brevity. The coordinates of the end effector are used to train a CNMP and a CNEP model. Conditioned on an observation (shown with $\bullet$), {\shortName} (shown in \textcolor{orange}{orange}) chooses one of the modes and generates motion trajectories closer to the demonstrations while {\otherShortName} (shown in \textcolor{blue}{blue}) interpolates between modes, leading to collisions (shown with \textcolor{red}{\textbf{X}}) with the obstacle (shown in \textcolor{darkgray}{gray}). \label{fig:baxter_traj_gen}}
    
\end{figure}

\begin{figure}[htbp]
    \centering
    \subfloat[]{\includegraphics[width=0.49\columnwidth]{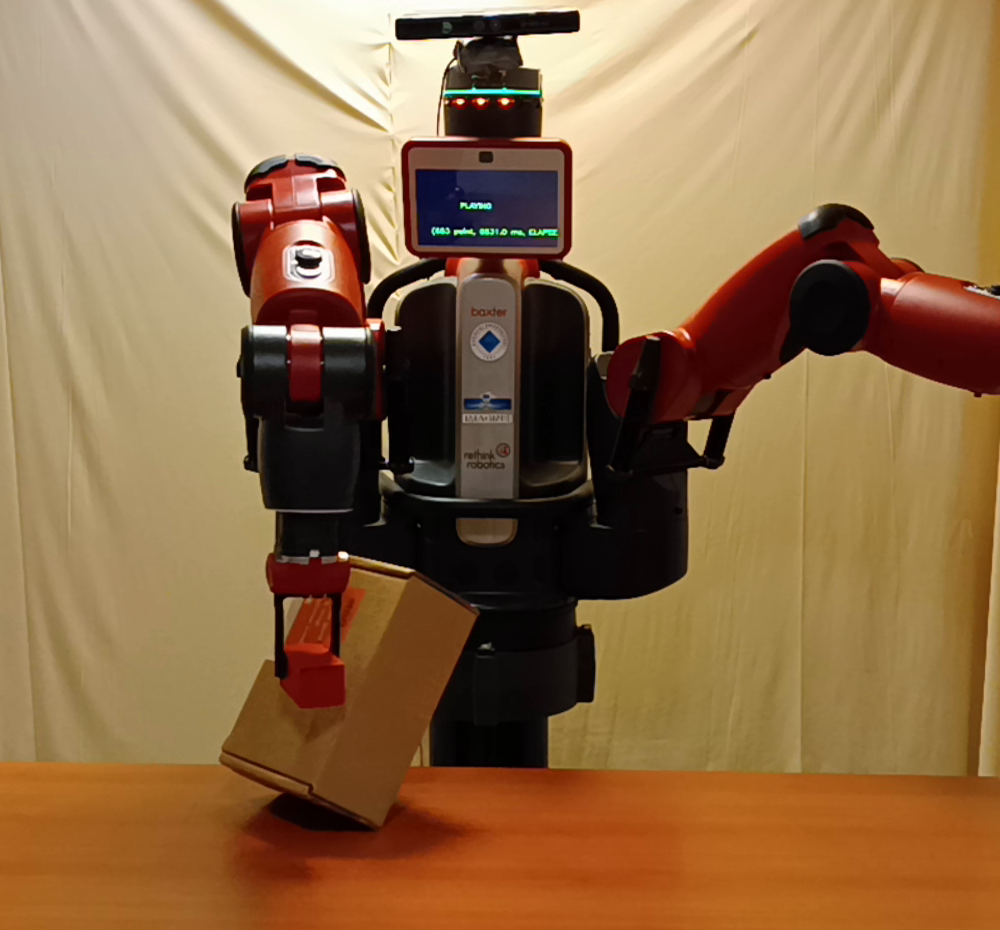}}
    \enspace
    \subfloat[]{\includegraphics[width=0.49\columnwidth]{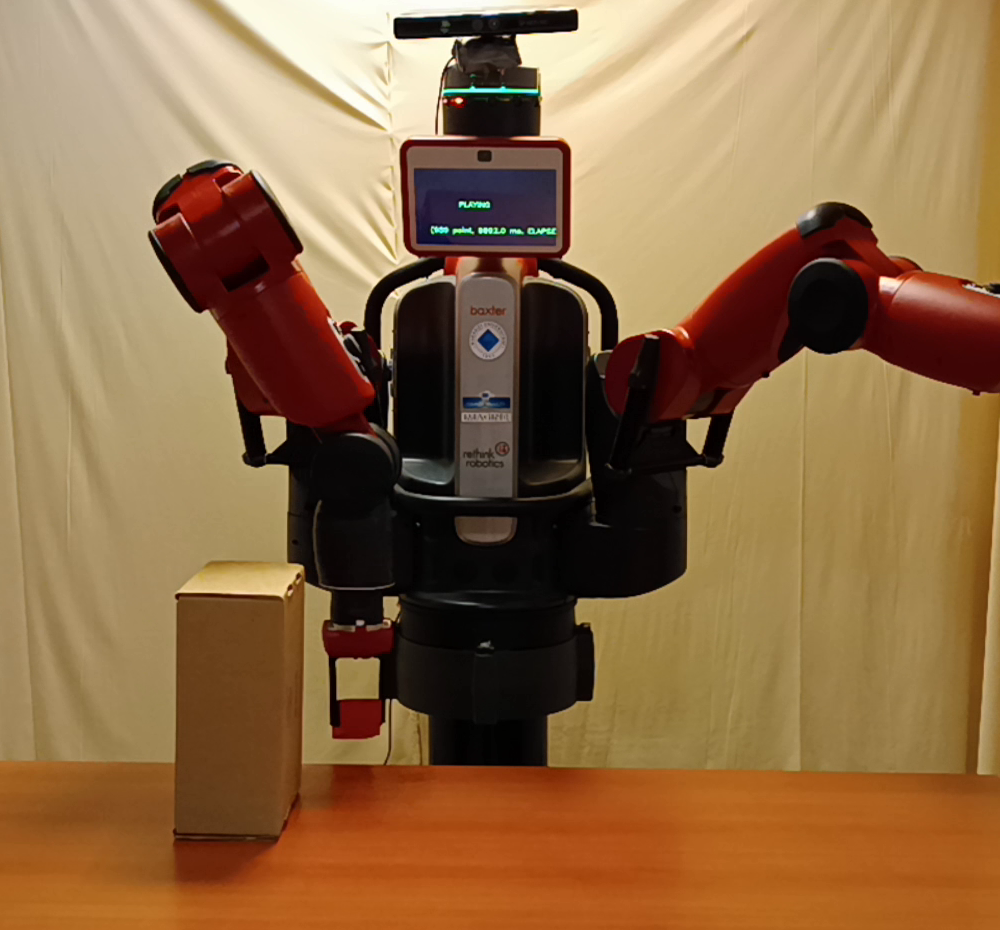}}
    \caption{In (a), the generated trajectory by the {\otherShortName} model collide with the box in the middle. In (b) and (d), {\shortName} only interpolates within demonstrations from one mode and, therefore, follows one of the possible modes, successfully generating paths that avoid obstacles.\label{fig:baxter_run}}
    
\end{figure}

After training, both models were requested to generate the obstacle avoidance skill. In the first case, where models were trained with the trajectories of end-effector positions, generated values were passed through a PID controller and an inverse kinematics module before the execution on the robot. Similarly, generated joint angle values are passed through a PID controller and a forward kinematics module in the second case for illustration purposes. Results in Fig. \ref{fig:baxter_traj_gen} indicate a distinctive pattern, supporting our initial claim. While the {\otherShortName} tends to interpolate between the provided demonstrations, the {\shortName} demonstrates a preference for adhering closely to the demonstrated behaviors. To explain the implications of this behavior for real-world robot tasks, we run the trained models on the real robot and present the results in Fig. \ref{fig:baxter_run}. Given these demonstrations, {\otherShortName}-generated trajectories lead to collisions while {\shortName}-generated trajectories can safely avoid obstacles.
The dataset from these demonstrations comprises two SM trajectories that move the robot arm from a start to an end position while avoiding an obstacle, and grasp the target object by pressing a button.

% COM1: removed to save space

\subsubsection{Pick-and-Place Wine Glasses on a Dish Rack}
\begin{figure}[t]
    \begin{center}
        \includegraphics[width=0.99\columnwidth]{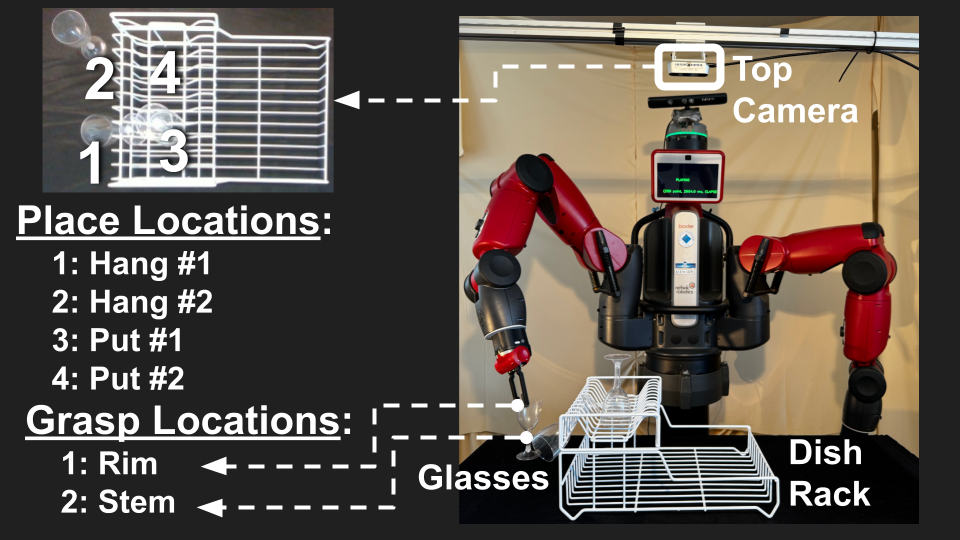}
        \caption{In this setup, the robot has access to an external camera (top camera) to pick wine glasses from the table top and place them on a dish drying rack. The top camera takes RGB pictures of the tabletop, which are used to extract spatial features of the objects. A wine glass offers 2 grasping locations for Baxter: (1) from the rim and (2) from the stem. Also, there are 4 placement options for each wine glass. Refer to the text for the details about the scenario.} \label{fig:bax_hd_setup}
        
    \end{center}
\end{figure}

\paragraph{Individual Skills}
In this experimental setup (Fig. \ref{fig:bax_hd_setup}), the robot is expected to pick wine glasses from the tabletop and place them on a designated dish rack. Each glass can be grasped from 2 different locations: (1) from the rim and (2) from the stem. If the robot grasps the wine glass from the rim, it can hang the glass to the side of the dish rack (Places 1 and 2 in Fig. \ref{fig:bax_hd_setup}). If it grasps from the stem, it can put the glass upside down on the top of the dish rack (Places 3 and 4).

Initially, the robot captures RGB images of the tabletop that has a resolution of 640x480 pixels. These images are then processed by the pre-trained object detection network MobileNet-v2 \cite{sandler2018mobilenetv2}. In this process, we exclude the classification layer at the end of the MobileNet-v2 model to focus on extracting only the spatial features of the objects in the image. This extraction results in a 1280-dimensional feature array that represents the spatial characteristics of the detected objects. Next, trajectories of target skills are demonstrated by an expert. Extracted feature arrays are combined with demonstration trajectories to form the input for CNEP. The complete input for the CNEP model is a list of 1288-dimensional trajectories: the 1280-dimensional feature array from MobileNet-v2, the 7-dimensional pose of the end-effector, and the status of the gripper (whether it is open or closed). Thus, the system integrates computer vision through MobileNet-v2 with the demonstration trajectories to control the robot executing this complex manipulation task.

\begin{figure}[t]
    \centering
    \subfloat{\includegraphics[width=0.99\columnwidth]{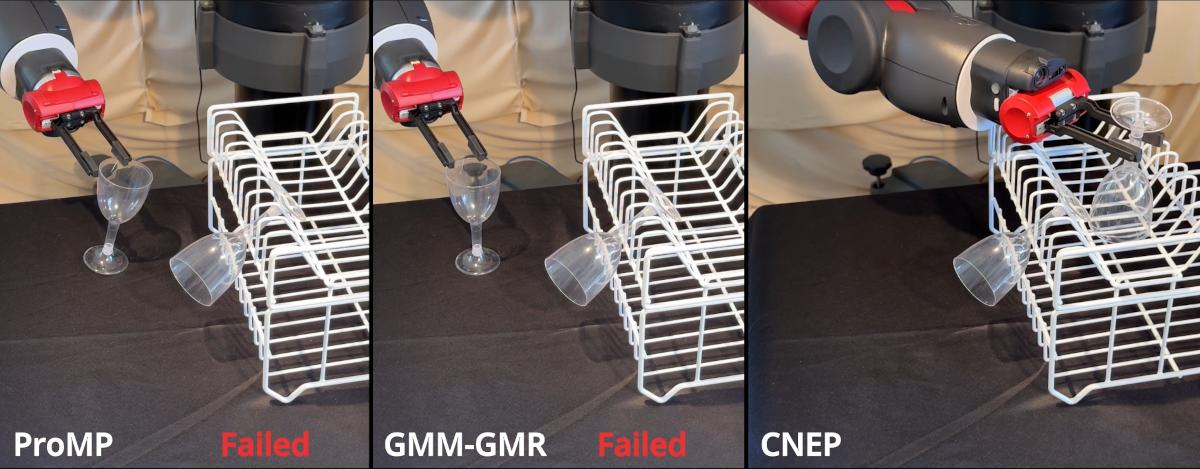}}
    \caption{40 pick and place demonstrations are provided. 3 models were trained on this data: a ProMP, a GMM, and a CNEP. Then, the wine glass was placed at a new location, and the models were conditioned on this observation. While ProMP and GMM failed to grasp the glass properly, CNEP successfully completed the task.}\label{fig:cnep_vs_pro_gmm}
    
\end{figure}

After training the CNEP model, we trained a ProMP and a GMM using the same input to compare the trajectory generation performances. When the data contained 1288 dimensions, both approaches failed to model trajectories. Therefore, the dimensionality of the data is reduced following the technique given in \cite{dermy2018prediction}. When testing, ProMP and GMM failed to grasp the glass, while CNEP successfully grasped, picked, and placed it, as shown in Fig. \ref{fig:cnep_vs_pro_gmm}. All demonstration and execution videos are available at \url{https://www.youtube.com/playlist?list=PLXWw0F-8m_ZZD7fpGOKclzVJONXUifDiY}.

\paragraph{Online Conditioning and On-the-fly Adaptation}

\begin{figure}[t]
    \centering
    \subfloat[]{\includegraphics[width=0.32\columnwidth]{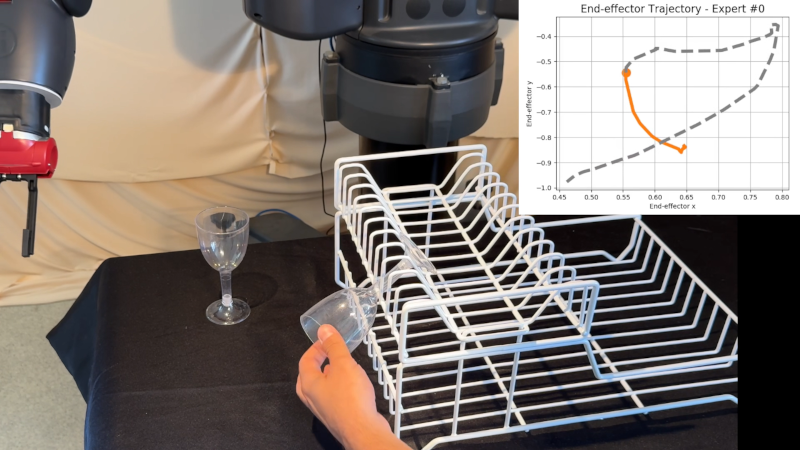}}
    \enspace
    \subfloat[]{\includegraphics[width=0.32\columnwidth]{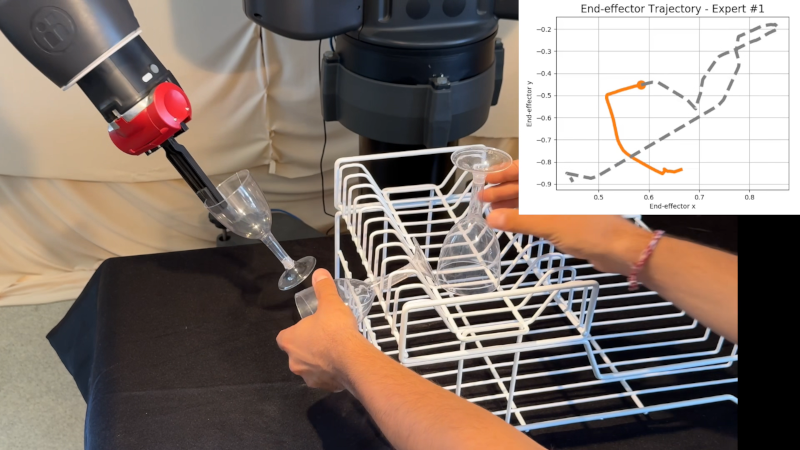}}
    \enspace
    \subfloat[]{\includegraphics[width=0.32\columnwidth]{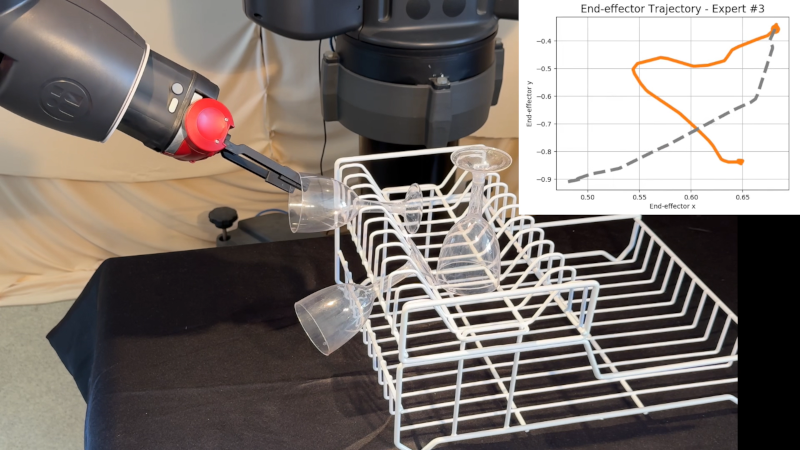}}
    \caption{When the tabletop configuration is changed during the execution of a trajectory, CNEP can adapt by continuously conditioning on the current sensory data and switching among experts on-the-fly. (a) CNEP chooses expert-0 to grasp the glass from the stem and place it straight on the dish rack. During execution, the hanged glass is removed. (b) CNEP switches to expert-1 to grasp the glass from the rim and hang it on the side of the dish rack. As it proceeds, the configuration changes again, and CNEP switches to expert-3, which hangs the glass to the available spot.} \label{fig:online}
    
\end{figure}

Conditioned on the real-time input from the camera, the position of the end-effector, and the status of the gripper, CNEP can produce real-time control commands to realize target skills.  As a proof-of-concept demonstration, in the same setup, we changed the tabletop configuration twice during the execution of the trajectory generated by the CNEP. Our system successfully reacted to the configuration changes by changing the responsible expert and, hence, the produced control commands, as shown in Fig. \ref{fig:online}.

\section{Conclusion \label{sec:con}}

In this study, we introduced an LfD method, namely the {\longName}. CNEP is proposed to improve the modeling and generation capabilities of LfD systems when available demonstrations correspond to diverse, multimodal sensorimotor trajectories. This is achieved by the utilization of (1) the novel architectural components, the Gate Network, and the experts, and (2) the novel components of the loss function, the \textit{batch entropy} and the \textit{individual entropy}.

Our experiments demonstrated that CNEP is a robust LfD approach for modeling and generating robotic skills even in real time. It successfully models intersecting multimodal or significantly different trajectories. It has better performance than baseline methods and effectiveness in real robot experiments. The number of experts is set manually and can be optimized as a hyper-parameter in the future. One limitation of the CNEP model is that, as a probabilistic framework, it does not guarantee passing through the observation points and requires using a higher-level module, such as a PID controller, to guarantee precision at observation points. Additionally, similar to other neural network-based approaches, the CNEP cannot extrapolate outside the training range.

\section*{Acknowledgment}
This research was funded by the European Union under the INVERSE project (101136067). The authors thank A. Ahmetoglu and I. Lirussi for their constructive feedback. The source code is available at \url{https://github.com/yildirimyigit/cnep}. 

%%%%%%%%%%%%%%%%%%%%%%%%%%%%%%%%%%%%%%%%%%%%%%%%%%%%%%%%%%%%%%%%%%%%%%%%%%%%%%%%%%%%%%%%
%%%%%%%%%%%%%%%%%%%%%%%%%%%%%%%%%%%%%%%%%%%%%%%%%%%%%%%%%%%%%%%%%%%%%%%%%%%%%%%%%%%%%%%%
%%%%%%%%%%%%%%%%%%%%%%%%%%%%%%%%%%%%%%%%%%%%%%%%%%%%%%%%%%%%%%%%%%%%%%%%%%%%%%%%%%%%%%%%

%\addtolength{\textheight}{-11.5cm}
\bibliographystyle{IEEEtran}
\bibliography{bibtex/bib/ref}

\end{document}